\documentclass[runningheads]{llncs}

\usepackage{_sty/eccv}

\usepackage{_sty/eccvabbrv}

\usepackage{booktabs}
\usepackage{graphicx}
\usepackage{multirow}
\usepackage{diagbox}
\usepackage{makecell}
\usepackage{pifont}
\usepackage{amsfonts}
\usepackage{wrapfig}
\usepackage{amsmath}
\usepackage{extarrows}
\usepackage{caption}
\usepackage{float}
\usepackage{algorithm}
\usepackage{algorithmic}
\usepackage{color}
\usepackage[table]{xcolor}

\usepackage[accsupp]{axessibility}  

\usepackage{etoolbox} 

\newbool{inccomment}
\booltrue{inccomment}  

\definecolor{mygray}{RGB}{230, 230, 230}

\newcommand{\XC}[1]{\ifbool{inccomment}{{\color{magenta}XC\@: #1}}{}}
\newcommand{\JY}[1]{\ifbool{inccomment}{{\color{blue}JY\@: #1}}{}}

\usepackage{hyperref}

\usepackage{orcidlink}

\begin{document}

\title{EcoVideo: Entropy-Orchestrated Video Generation Paradigm in Cloud-Edge Dynamics} 

\titlerunning{EcoVideo}


\author{
Jiayu Chen\inst{1}\textsuperscript{*}\orcidlink{0009-0002-5428-8647}
\and Hengyi Zhang\inst{3}\textsuperscript{*}\orcidlink{0009-0005-8512-9894}
\and Maoliang Li\inst{1}\orcidlink{0009-0008-6773-5067}
\and Minyu Li\inst{4}\orcidlink{0009-0006-9567-2353}
\and Zihao Zheng\inst{1}\orcidlink{0009-0008-4624-2853} 
\and Xuanzhe Liu\inst{1}\orcidlink{0000-0002-7908-8484}
\and Guojie Luo\inst{1, 2}\orcidlink{0000-0003-4932-3655}
\and Xiang Chen\inst{1, 2}\textsuperscript{$\dagger$}\orcidlink{0000-0003-2790-976X}
}

\authorrunning{~Jiayu Chen et al.}

\institute{School of Computer Science, Peking University, China
\and State Key Lab of Multimedia Information Processing, Peking University, China
\and International School, Beijing University of Posts and Telecommunications, China
\and School of Information Science and Technology, Beijing Forestry University, China
\\[0.5em] \textsuperscript{*}Equal contribution. \quad \textsuperscript{$\dagger$}Corresponding author: xiang.chen@pku.edu.cn
}

\maketitle


\begin{abstract}
    DiT video generation is latency-intensive due to iterative full-frame denoising, while prior cloud-edge methods largely rely on static inter-step decoupling and cannot leverage inter-frame similarity or adapt to system dynamics. We propose \textbf{EcoVideo}, an entropy-orchestrated framework for \emph{dynamic inter-frame decoupling}: early-stage self-attention entropy provides a training-free estimate of frame-wise information density for frame selection; a cloud large model denoises sparse high-entropy keyframes; and an edge lightweight model reconstructs the remaining frames via motion-aware interpolation with refinement for temporal stability. EcoVideo further adapts the keyframe budget and edge refinement depth to real-time bandwidth and compute availability, optimizing end-to-end latency under constraints. Experiments on representative DiT video generators show improved quality--efficiency trade-offs and up to \textbf{2.9$\times$} end-to-end speedup in low-bandwidth, compute-limited edge settings. Code is available at \url{https://github.com/IF-LAB-PKU/EcoVideo}.
\end{abstract}

\begin{figure}[H]
    \includegraphics[width=\linewidth]{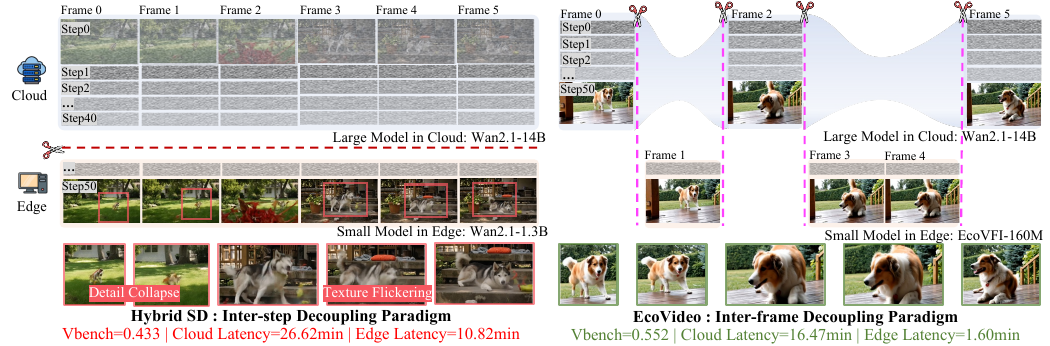}
        \vspace{-6mm}
        \caption{A comparison between inter-step decoupling (HybridSD) and inter-frame decoupling paradigms (ours).}
    \label{fig:1}
\end{figure}

\section{Introduction}

In recent years, Diffusion Transformer (DiT)\cite{vaswani2017attention,peebles2023scalable} based video generation models\cite{wan2025wan,yang2024cogvideox} have achieved remarkable progress in visual fidelity and spatiotemporal coherence.
    However, the DiT paradigm is inherently built upon step-by-step full-frame denoising\cite{ho2020denoising}, which introduces a severe computational bottleneck and makes large-scale video generation services difficult to deploy in practice.

To relieve the computation pressure, recent studies\cite{pan2025tstitch,cheng2025srdiffusion,yan2024hybrid,xie2025ec} have started to explore a cloud-edge collaborative generation paradigm.
    Representative works such as HybridSD\cite{yan2024hybrid} and EC-Diff\cite{xie2025ec} attempt to decouple inter-step dependencies in generation: a large model on the cloud performs early denoising steps to establish the semantic layout, while a small model on the edge completes the remaining refinement steps.
    Although promising, existing \textit{inter-step decoupling} paradigms exhibit intrinsic limitations in practice, as shown in the left of Fig.~\ref{fig:1}.
        Algorithmically, the small edge model often lacks sufficient denoising capability. Due to the limited denoising capacity of the small edge model, the late-stage denoising tends to overly smooth the video, often leading to informative frame artifacts such as texture flickering and detail collapse.
        More critically, inter-step decoupling implicitly assumes that every frame is equally important. Even when adjacent frames are highly similar, the pipeline still repeats nearly identical denoising processes frame by frame, leading to substantial redundant computation along the temporal dimension.
        Systemically, real-world cloud-edge environments are highly dynamic: fluctuations in network bandwidth and variations in cloud and edge resources cause the optimal workload partition to change over time. Therefore, a static \textit{inter-step decoupling} paradigm cannot adapt, resulting in unstable or even degraded end-to-end efficiency in practice.

Furthermore, we argue that the above issues provide three key insights into optimizing decoupled video generation under dynamic cloud–edge environments, as shown in the right of Fig.~\ref{fig:1}.
\textbf{Insight 1: }\textit{Inter-step decoupling overlooks inter-frame sparsity, leading to inefficient generation.}
    Unlike \textit{inter-step decoupling}, which still produces many similar and redundant frames, one must identify inter-frame differences and sparse information density along the temporal dimension. The decoupling paradigm should online localize keyframes that capture critical motion changes or structural transitions, thereby avoiding redundant per-frame generation in the cloud.
\textbf{Insight 2:}\textit{Inter-frame decoupling video generation can be enabled by information reorganization and reuse.}
    When the large cloud model generates only some frames to compress workload, the small edge model synthesis of intermediate frames \cite{danier2024ldmvfi,zhang2025eden} must reuse generated frame-conditioned cues to minimize temporal information loss between the edge and cloud outputs, preserving motion trajectories, structural boundaries, and texture details, and preventing flicker or abrupt temporal detail shifts.
\textbf{Insight 3: } \textit{Inter-frame decoupling need adapt to cloud–edge dynamics.}
    Given time-varying system conditions (e.g., network bandwidth and compute capability), the decoupling paradigm should adaptively allocate generative workloads across the cloud and edge. This enables closed-loop, end-to-end optimization under dynamic cloud-edge environments.

\begin{figure*}[t]
    \includegraphics[width=\linewidth]{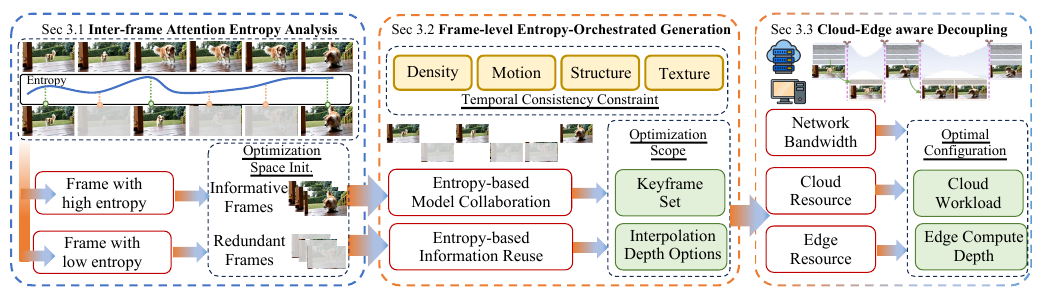}
    \caption{Overview of EcoVideo. \textbf{Sec \ref{sec:3.1}:} Inter-frame attention-entropy analysis estimates information density. \textbf{Sec \ref{sec:3.2}:} Frame-level entropy-orchestrated generation preserves temporal consistency via model collaboration and information reuse. \textbf{Sec \ref{sec:3.3}:} Cloud-edge dynamics-aware decoupling adaptation searches the optimal configuration.}
    \label{fig:2}
\end{figure*}

Building on these insights, we propose an entropy-orchestrated dynamic cloud-edge video generation paradigm, termed \textbf{EcoVideo}.
    The key idea is to shift the cloud-edge collaboration perspective from \emph{static inter-step decoupling} to \emph{dynamic inter-frame decoupling and orchestration}.
    Specifically, as shown in Fig.~\ref{fig:2} \textbf{guided by Insight 1}, we use attention entropy as a training-free signal to estimate online frame-wise information density and separate informative frames from low-entropy, highly similar frames, thereby initializing an entropy-based optimization space. Intuitively, lower entropy typically reflects high similarity and temporal redundancy.
    \textbf{Guided by Insight 2}, the decoupling is centered on bounding the optimization scope under quality constraints. EcoVideo orchestrates generation in two collaborative stages: \textit{Inter-frame Decoupling and Model Collaboration}, and \textit{Inter-frame Information Reuse and Interpolation}. Concretely, the large cloud model denoises only a small set of high-entropy informative frames to anchor the global motion skeleton. Meanwhile, it injects frozen cues from similar frames as temporal-detail context, helping preserve motion trajectories and overall structure. The small edge model then reuses these frame-conditioned cues to synthesize the remaining low-entropy intermediate frames via motion-aware interpolation with greedy refinement, minimizing temporal information loss and effectively mitigating flicker and abrupt temporal-detail drift.
    \textbf{Guided by Insight 3}, we further develop a \textit{cloud–edge dynamics-aware decoupling adaptation} mechanism. By searching the optimization scope for the optimal cloud workload and edge compute depth based on network bandwidth and cloud-edge resources, EcoVideo balances cloud-side workload and end-to-end latency while maintaining runtime temporal coherence.

We conduct systematic evaluations of EcoVideo on mainstream DiT model families, including Wan2.1-T2V\cite{wan2025wan}, Wan2.2-T2V, and CogVideoX\cite{yang2024cogvideox}, using VBench\cite{zheng2025vbench} as the primary metric. Experimental results demonstrate that, compared with HybridSD and EC-Diff, EcoVideo significantly reduces cloud computation and transmission costs, while effectively mitigating temporal detail discontinuities commonly observed in existing cloud-edge methods, such as texture flickering and detail collapse. In low-bandwidth and compute-limited edge scenarios, EcoVideo achieves up to \textbf{2.9$\times$} end-to-end speedup.
\paragraph{Contributions.}
Our main contributions are threefold:
\begin{itemize}
    \item To our knowledge, we are the first to introduce an \textbf{inter-frame decoupling paradigm} into cloud-edge video generation, shifting large-small model collaboration from inter-step model switching to frame-level workload partitioning that exploits temporal redundancy.
    \item We design a \textbf{training-free inter-frame attention-entropy analysis} module that estimates frame-wise information density and identifies high-information keyframes for cloud-side generation.
    \item We propose a \textbf{frame-level entropy-orchestrated generation framework} that preserves temporal consistency through model collaboration and information reuse, and adapts the keyframe budget and edge refinement depth to cloud--edge dynamics, achieving up to \textbf{2.9$\times$} end-to-end speedup.
\end{itemize}

\section{Preliminary and Problem Definition}
\label{sec:preliminary}


\subsection{Video Diffusion Generation Paradigm}
\label{sec:2.1}
Video diffusion models follow a step-by-step full-frame denoising paradigm.
    During inference, given a conditioning input $c$, the model first initializes a length-$F$ noise sequence in the video latent space,
    $\mathbf{x}_T \in \mathbb{R}^{F\times H\times W\times C}$, where each latent feature map corresponds to fixed video frames, and $\mathbf{x}_T \sim \mathcal{N}(\mathbf{0}, \mathbf{I})$.
    The diffusion model then refines $\mathbf{x}_T$ into the target latent $\mathbf{x}_0$ through a pre-defined number of denoising steps $T$, and finally decodes it into a pixel-space video via a VAE.
    \begin{equation}
    \text{Denoising process: }
    \mathbf{x}_{t-1} = f(t)\mathbf{x}_t - g(t)\epsilon_\theta(\mathbf{x}_t,t,c), \quad t=1,\ldots,T,
    \end{equation}
    where $\epsilon_{\theta(\cdot)}$ is the noise-prediction network (e.g., VideoDiT) that jointly models cross-frame dependencies along both spatial and temporal dimensions, and $f(t), g(t)$ are sampler-specific coefficients.
Since each denoising step requires a full prediction over the $F$-frame sequence, the overall computation increases sharply with both the denoising steps $T$ and the sequence length $F$, making large-scale video generation services difficult to deploy.

Some recent works\cite{yan2024hybrid,xie2025ec} attempt to accelerate inference by \textit{inter-step decoupling} and collaboration.
Specifically, a large video DiT model $\epsilon_{\theta_1}$ performs early timesteps $t\in(T,\ldots,t_1]$, and a smaller video DiT model $\epsilon_{\theta_2}$ takes over the remaining steps $t\in(t_1,\ldots,1]$.
    However, this switch changes the noise estimates along the sampling trajectory.
        Let the prediction gap after switching be
        $\Delta\epsilon_t=\epsilon_{\theta_2}(\mathbf{x}_t,t,c)-\epsilon_{\theta_1}(\mathbf{x}_t,t,c)$.
        In practice, $\Delta\epsilon_t$ is often inconsistent across frames, which amplifies inter-frame discrepancies and breaks temporal coherence.
    Consequently, high-frequency textures and cross-frame consistency become difficult to converge, manifesting as temporal artifacts such as texture flickering.

Motivated by this issue, we argue that the decoupling perspective should shift from \textit{inter-step decoupling} to \textit{inter-frame decoupling}:
    One should identify information-density differences along the temporal axis, let the large model generate keyframes that carry critical motion or structural transitions, 
    and let the lightweight model complete intermediate frames, thereby avoiding temporal inconsistency induced by the capability gap between large and small models.

\subsection{Cloud--Edge Generation Paradigm}
\label{sec:2.2}
Cloud-edge collaboration offloads part of the generation workload to the edge, reducing cloud-side load and improving service throughput.
    In a typical setup, a large model runs on the cloud while a lightweight model runs on the edge, and the end-to-end latency can be decomposed as
    $L = L_{\text{cloud}} + L_{\text{net}} + L_{\text{edge}}$.
    According to previous works\cite{yang2024efficient,wang2024cloud,yan2024hybrid}, $L_{\text{cloud}}$ and $L_{\text{edge}}$ depend on the available compute resources and the allocated workloads on the cloud and edge, respectively, while $L_{\text{net}}$ is determined by the bandwidth and the amount of transmitted data.

Taking the \textit{inter-step decoupling} paradigm as an example, suppose the cloud executes the first $t_1$ steps and the edge executes the remaining $T-t_1$ steps, while transmitting intermediate results of size $S(t_1)$.
    The total latency can be written as
    \begin{equation}
    L(t_1;B,P_c,P_e)=\underbrace{\frac{C_{\text{cloud}}(t_1)}{P_c}}_{L_{\text{cloud}}}
    +\underbrace{\frac{S(t_1)}{B}}_{L_{\text{net}}}
    +\underbrace{\frac{C_{\text{edge}}(T-t_1)}{P_e}}_{L_{\text{edge}}},
    \end{equation}
    where $B$ is the instantaneous bandwidth, $P_c$ and $P_e$ denote the available cloud/edge compute, and $C_{\text{cloud}}$ and $C_{\text{edge}}$ are the corresponding computational loads.
Since environmental parameters such as $B$ and $P_e$ can fluctuate significantly over time, a fixed splitting point $t_1$ cannot remain optimal.
    When bandwidth drops or edge compute becomes contended, the communication term $S(t_1)/B$ and the edge-side computation term $C_{\text{edge}}(T-t_1)/P_e$ can jointly dominate the end-to-end latency, leading to overall degradation:
    \begin{equation}
    L(t_1;B,P_c,P_e) > L_{\text{all-cloud}}=\frac{C_{\text{cloud}}(T)}{P_c}.
    \end{equation}
This implies that fixed step splitting may not only fail to accelerate under limited bandwidth or edge compute, but also incur extra communication and edge bottlenecks, causing unstable or degraded end-to-end latency.

Therefore, it is necessary to establish an optimal mapping between the video generation paradigm and the dynamic cloud–edge system, enabling resource-aware end-to-end optimization under time-varying conditions.

\section{Method}
\label{sec:method}

In this section, we introduce \textbf{EcoVideo}, an entropy-orchestrated cloud-edge collaborative paradigm for efficient and robust video generation. 
In contrast to static \textit{inter-step decoupling}, EcoVideo reformulates collaboration as dynamic \textit{inter-frame decoupling and orchestration}. 
Figure~\ref{fig:2} summarizes the overall workflow. 
    EcoVideo comprises three components:
    (i) \textbf{Frame-level attention entropy analysis} computes token-wise attention entropy and aggregates it into a stable frame-entropy sequence, serving as a training-free proxy to estimate online frame-wise information density and initialize an entropy-based optimization space.
    (ii) Based on this signal, \textbf{frame-level entropy-orchestrated generation} determines the optimization scope under quality constraints by selecting informative keyframes for cloud–edge collaboration, where the cloud-side large model generates high-entropy keyframes and the edge-side small model synthesizes the remaining frames via keyframe-conditioned interpolation and information reuse, thereby forming a dynamic configuration space over keyframe selections and edge interpolation depths.
    (iii) \textbf{Cloud-edge dynamics-aware decoupling adaptation} searches the optimization scope to adapt optimal orchestration configuration to real-time bandwidth and compute resources, enabling reliable performance under cloud-edge dynamics.

\subsection{Inter-frame Attention Entropy Analysis}
\label{sec:3.1}  

As discussed in Section \ref{sec:2.1}, existing \textit{inter-step decoupling} methods can introduce fine-grained temporal artifacts due to the capability gap between small and large models. 
    To address this issue, EcoVideo adopts inter-frame decoupling rather than inter-step decoupling: 
        It recognizes the non-uniform information density along the temporal dimension, 
        and decomposes the video generation paradigm into keyframe generation with high dynamics and high uncertainty, and intermediate-frame interpolation for low-dynamic and stable regions.

Inspired by prior work\cite{chentoprovar,ma2025towards} that uses attention entropy to measure information distribution and uncertainty, we aggregate token-level attention entropy to the frame level to evaluate the temporal information density of different frames.
    Concretely, at diffusion step $t$, let the self-attention weight matrix be $\mathbf{A}^{(t)} \in [0,1]^{N\times N}$. 
    For each query token $i$, we define its attention entropy as $H^{(t)}_i = -\sum_{j=1}^{N} A^{(t)}_{i,j}\log\!\left(A^{(t)}_{i,j}\right)$. 
    We then aggregate token entropies to the frame level according to the frame index mapping $\pi(i)=f$: 
\begin{equation}
e^{(t)}_f= \mathcal{P}\big(\{\,H^{(t)}_i \mid \pi(i)=f\,\}\big),
\label{eq:frame_entropy_pool}
\end{equation}
    where $\mathcal{P}(\cdot)$ is an aggregation operator. To enhance robustness, we adopt mean aggregation:
\begin{equation}
e^{(t)}_f = \frac{1}{|\mathcal{I}_f|}\sum_{i\in\mathcal{I}_f} H^{(t)}_i,
\qquad
\mathcal{I}_f = \{\, i \mid \pi(i)=f \,\}.
\label{eq:frame_entropy_mean}
\end{equation}
    A larger frame-level entropy indicates more dispersed attention, more complex dynamics, and higher information density. 
    We therefore treat high-entropy frames as keyframe candidates, prioritizing them to be modeled by the generative model, while leaving low-entropy frames to be completed via interpolation.

Furthermore, since attention patterns are more unstable in early denoising steps, consistent with prior work\cite{xi2025sparse,zhang2025fast}, EcoVideo defines the first $10\%$ denoising steps as a warm-up stage. 
    EcoVideo computes $e^{(t)}_f$ only during this stage, and fuses estimates across steps using an exponential moving average (EMA) to obtain a stable estimate:
\begin{equation}
\tilde e^{(t)}_f = \alpha e^{(t)}_f + (1-\alpha)\tilde e^{(t-1)}_f,
\qquad
e_f = \tilde e^{(t_{\mathrm{w}})}_f,
\label{eq:frame_entropy_ema}
\end{equation}
    and outputs the final stable frame-level entropy as $e_f$, where $t_{\mathrm{w}}$ denotes the last step of the warm-up stage. Finally, we use the stable frame-level entropy $\{e_f\}$ obtained from the warm-up stage as the criterion for keyframe selection.

\begin{figure*}[t]
    \includegraphics[width=\linewidth]{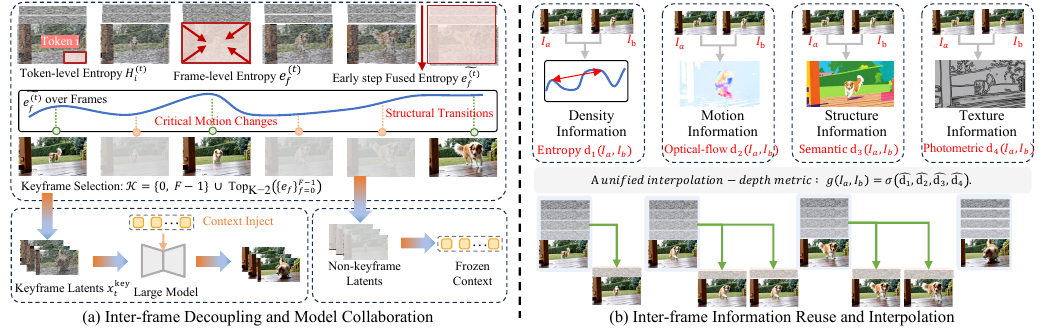}
    \caption{
    Frame-level entropy-orchestrated generation. 
    (a) Inter-frame decoupling and model collaboration select high-information keyframes and inject frozen non-keyframe context for cloud-side denoising. 
    (b) Inter-frame information reuse and interpolation fuse density, motion, structure, and texture cues to guide edge-side refinement.
}
    \label{fig:4}
\end{figure*}

\subsection{Frame-level Entropy-Orchestrated Generation}
\label{sec:3.2}

Although attention entropy provides a training-free proxy to differentiate frame-level information density, EcoVideo must further address two key challenges to preserve fidelity under collaborative splitting:
    (i) how to denoise only keyframes while avoiding quality degradation caused by losing global temporal layout; and
    (ii) how to ensure that edge-completed intermediate frames remain spatiotemporally consistent with cloud-generated keyframes, especially around structural boundaries, motion trajectories, and fine textures.

\noindent{\textbf{Entropy-based Inter-frame Decoupling and Model Collaboration.}}
To reduce temporal inconsistency, EcoVideo enforces the first and last frames to be keyframes and then selects the remaining keyframes by entropy ranking.
    Specifically, given the stable entropy sequence $\{e_f\}_{f=0}^{F-1}$, we choose the keyframe index set as
    $\mathcal{K}=\{0,\;F-1\}\ \cup\ \mathrm{Top}_{K-2}\Big(\{e_f\}_{f=0}^{F-1}\Big)$,
    where $\mathrm{Top}_{K-2}(\cdot)$ returns the indices of the largest $K-2$ values, and $|\mathcal{K}|=K$.
    After sorting $\mathcal{K}$ in chronological order, we obtain $(k_0<k_1<\cdots<k_{K-1})$.

Let $\mathbf{x}_{t_{\mathrm{w}}}\in\mathbb{R}^{F\times H\times W\times C}$ denote the latent at the end of the warm-up.
EcoVideo performs latent cropping along the temporal axis and keeps only the keyframe slices:
    $\mathbf{x}^{\text{key}}_{t_{\mathrm{w}}}=\mathcal{C}_{\mathcal{K}}\!\left(\mathbf{x}_{t_{\mathrm{w}}}\right)$, $\mathbf{x}^{\text{non}}_{t_{\mathrm{w}}}=\mathcal{C}_{\overline{\mathcal{K}}}\!\left(\mathbf{x}_{t_{\mathrm{w}}}\right)$,
    where $\mathcal{C}_{\mathcal{K}}(\cdot)$ extracts temporal slices indexed by $\mathcal{K}$, and $\overline{\mathcal{K}}$ denotes the complement index set.
The large model $\epsilon_{\theta_c}$ then continues denoising only on $\mathbf{x}^{\text{key}}_t$ for the remaining steps $t=t_{\mathrm{w}},\ldots,1$:
\begin{equation}
\mathbf{x}^{\text{key}}_{t-1}
=f(t)\mathbf{x}^{\text{key}}_{t}-g(t)\,\epsilon_{\theta_c}\!\left(\mathbf{x}^{\text{key}}_{t},t,c\right),
\qquad t=t_{\mathrm{w}},\ldots,1.
\label{eq:key_denoise}
\end{equation}
    
A naive keyframe-only denoising, however, can lose awareness of the global temporal layout encoded in the full-$F$ sequence, harming motion consistency and structure.
To mitigate this, EcoVideo injects global context by treating non-keyframe latents as frozen context tokens throughout the subsequent attention computations.
Concretely, we tokenize latents and condition inputs, and use non-keyframe tokens as additional keys/values with stop-gradient:
\begin{equation}
\epsilon_{\theta_c}\!\left(\mathbf{x}^{\text{key}}_{t},t,c\right)
=
\Big[\,
\epsilon_{\theta_c}\big(
\mathrm{Tok}(\mathbf{x}^{\text{key}}_{t}),
\ \mathrm{Tok}(\mathrm{sg}(\mathbf{x}^{\text{non}}_{t_{\mathrm{w}}})),
\ \mathrm{Tok}(c)
\big)
\,\Big]_{\mathcal{K}},
\label{eq:context_inject}
\end{equation}
where $\mathrm{Tok}(\cdot)$ denotes tokenization, $\mathrm{sg}(\cdot)$ stops gradients, and $[\cdot]_{\mathcal{K}}$ indicates that we only keep the noise prediction at keyframe positions.
Intuitively, $\mathbf{x}^{\text{non}}_{t_{\mathrm{w}}}$ provides a lightweight yet global temporal scaffold, allowing keyframe denoising to retain temporal context while still enjoying the speedup from temporal cropping.

\noindent{\textbf{Information Reuse and Inter-frame Interpolation.}}
After the cloud produces keyframe pixels $\mathcal{F}_{\mathcal{K}}=\{I_{0},I_{1},\ldots,I_k\}$, the edge completes the intermediate frames using a lightweight interpolation model, denoted as $\mathrm{EcoVFI}(\cdot,\cdot)$.
    A central difficulty is that uniform interpolation can over-allocate budget to static segments while under-refining high-motion segments, leading to temporal artifacts such as abrupt changes, ghosting, or local blur.
EcoVideo therefore adopts an entropy-aware greedy refinement module, termed EcoVFI, to dynamically allocate interpolation steps to the most challenging temporal intervals.

EcoVFI modifies EDEN's~\cite{zhang2025eden} interval scoring to fit our keyframe and edge refinement paradigm and to better preserve temporal consistency.
For each adjacent keyframe interval $(I_a,I_b)$, we compute a difficulty score $g(I_a,I_b)\in(0,1)$ to prioritize edge refinement. The score identifies intervals where interpolation is likely to fail, such as fast motion, structural changes, appearance shifts, or large entropy variation. We use four complementary cues, normalized per video as $\hat d_i\in[0,1]$: motion $d_1$ from optical-flow magnitude, density $d_2$ from the keyframe entropy gap, structure $d_3$ from endpoint feature distance, and texture $d_4$ from pixel-domain appearance difference.


{\small
\noindent
\begin{minipage}[t]{0.48\linewidth}
\begin{equation}\label{eq:cue_motion}
d_1(I_a,I_b)=\frac{1}{|\Omega|}\sum_{p\in\Omega}\|u(p)\|_2
\end{equation}
\vspace{-10pt}
\begin{equation}\label{eq:cue_structure}
d_3(I_a,I_b)=\|\phi(I_a)-\phi(I_b)\|_2
\end{equation}
\end{minipage}\hfill
\begin{minipage}[t]{0.48\linewidth}
\begin{equation}\label{eq:cue_density}
d_2(I_a,I_b)=|e_a-e_b|
\end{equation}
\begin{equation}\label{eq:cue_texture}
d_4(I_a,I_b)=\frac{1}{|\Omega|}\|I_a-I_b\|_1
\end{equation}
\end{minipage}
}

where $e_a,e_b$ are the frame-level entropies of keyframes $I_a$ and $I_b$, and $\Omega$ denotes the pixel domain. 
We instantiate $\phi(\cdot)$ as DINO-small~\cite{caron2021emerging} for semantic structural feature extraction, and use RAFT~\cite{teed2020raft} to compute bidirectional optical flow between adjacent keyframes, with $u(p)$ denoting the flow vector at pixel $p$. 
We aggregate the four normalized cues into a single score and apply a sigmoid mapping to $(0,1)$:
\begin{equation}\label{eq:interval_score}
g(I_a,I_b)=\sigma\!\left(\frac{\hat d_1+\hat d_2+\hat d_3+\hat d_4}{4}\right).
\end{equation}
A higher score implies stronger motion, structure, or texture changes, and thus calls for deeper edge refinement to preserve temporal consistency.
This design assigns higher scores to intervals with stronger motion, lower flow confidence, greater occlusion, larger appearance gaps, and larger entropy discrepancies, exactly the cases where interpolation is prone to artifacts.

Overall, EcoVFI assigns each adjacent keyframe interval an initial difficulty score $s_i=g(I_i,I_{i+1})$ and greedily refines the sequence until reaching length $F$ by repeatedly selecting the highest-score interval $(I_a,I_b)$, inserting its midpoint $I_m=\mathrm{EcoVFI}(I_a,I_b)$ to split it into $(I_a,I_m)$ and $(I_m,I_b)$, and updating the corresponding scores, which concentrates interpolation budget on challenging regions and mitigates temporal discontinuities and ghosting.

\subsection{Cloud–Edge Dynamics-aware Decoupling Adaptation}
\label{sec:3.3}

Real-world cloud--edge environments are highly dynamic: fluctuations in bandwidth and variations in cloud/edge compute availability can change the optimal workload partition over time. 
To maintain stable end-to-end efficiency, EcoVideo includes a lightweight system-aware scheduler that maps real-time system states to two configuration knobs: the keyframe budget $K$ (i.e. keyframe density generated on the cloud) and the interpolation depth $D$ (i.e. refinement budget for edge interpolation). Intuitively, increasing $K$ provides stronger temporal anchoring but incurs higher cloud computation and transmission, while increasing $D$ improves intermediate-frame fidelity at the cost of additional edge computation. At runtime, the scheduler monitors instantaneous bandwidth $B$ and the available compute budgets on the cloud and edge, denoted by $P_c$ and $P_e$. Given the target video length $F$ and candidate sets $\mathcal{S}_K$ and $\mathcal{S}_D$, it outputs $(K,D)$ for the next request (or scheduling window).

Following the standard cloud--edge decomposition, we model the end-to-end latency as an additive sum:
\begin{equation}
    T(K,D;B,P_c,P_e)=T_{\text{warm}} + T_{\text{cloud}}(K;P_c) + T_{\text{net}}(K;B) + T_{\text{edge}}(K,D;P_e),
\label{eq:dyn_latency_sum}
\end{equation}
where $T_{\text{warm}}$ is a fixed warm-up overhead (for entropy estimation and keyframe decision). Each term is estimated via online profiling. 
Concretely, we maintain three unit statistics updated with an exponential moving average (EMA): (i) cloud time per keyframe $\bar{c}_c$, (ii) transmitted payload per keyframe $\bar{s}$, and (iii) edge time per interpolated frame $\bar{c}_e(D)$, which is non-decreasing in $D$. This yields the following approximations:
\begin{equation}
    \widehat{T}_{\text{cloud}} = \frac{K\,\bar{c}_c}{P_c},
    \widehat{T}_{\text{net}} = \frac{K\,\bar{s}}{B},
    \widehat{T}_{\text{edge}} = \frac{(F-K)\,\bar{c}_e(D)}{P_e}.
\label{eq:dyn_cost}
\end{equation}

To select $(K,D)$ under resource constraints, we maximize a lightweight utility $Q(K,D)$ that captures diminishing returns, which is elaborated in the Appendix. 
We then solve a small discrete optimization:
\begin{equation}
    (K^*,D^*) = \arg\max_{K\in \mathcal{S}_K,\ D\in \mathcal{S}_D} \Big( Q(K,D) - \lambda\,\widehat{T}(K,D;B,P_c,P_e) \Big),
\label{eq:dyn_obj_sum}
\end{equation}
Here, $\lambda$ controls the quality-latency preference (e.g., set by a service-level target, or increased when the predicted latency exceeds a safety threshold). 
Since $\mathcal{S}_K$ and $\mathcal{S}_D$ are small, exhaustive search introduces negligible overhead.

\section{Experiments}

\subsection{Experimental Setup}

\textbf{Models and Baselines.}
We evaluate EcoVideo on a suite of open-source state-of-the-art DiT-based video generation models\cite{wan2025wan,yang2024cogvideox}, covering Wan2.1, Wan2.2, and CogVideoX.
Specifically, the cloud-side backbones are Wan2.1-14B, Wan2.2-A14B, and CogVideoX-5B, while the edge side uses our modified interpolation model EcoVFI-160M.
We compare EcoVideo against two representative cloud-edge collaboration baselines that rely on \emph{step decoupling}, namely HybridSD\cite{yan2024hybrid} and EC-Diff\cite{xie2025ec}.
For fairness, the cloud-side large model is kept identical across all methods.
For Wan backbones, both HybridSD and EC-Diff use Wan2.1-1.3B as the edge model; for CogVideoX, both use CogVideoX-2B.
All video-generation hyperparameters are aligned.

\noindent\textbf{Generation Settings.}
Wan2.1 and Wan2.2 generate 5-second videos with 81 frames at 720P resolution.
Wan2.1 uses 50 denoising steps, and Wan2.2 uses 40 denoising steps.
CogVideoX generates 6-second videos with 49 frames at 480P resolution using 50 steps.
More configurations are provided in the appendix.

\noindent\textbf{Datasets and Metrics.}
We use the VBench~\cite{zheng2025vbench} prompt set to assess both quality and efficiency.
For quality, we adopt VBench~\cite{zheng2025vbench} and report the overall score along with six representative dimensions—Subject Consistency, Background Consistency, Motion Smoothness, Aesthetic Quality, Imaging Quality, and Overall Consistency, to comprehensively measure fidelity and temporal coherence.
For efficiency, we report cloud latency, edge latency, communication volume, end-to-end latency, and speedup over the cloud-only baseline.

\noindent\textbf{Implementation Details.}
All experiments are conducted on a real cloud-edge system consisting of a single NVIDIA H200 (141GB) GPU and an NVIDIA Jetson Thor. 
    The average cloud-edge bandwidth is 20 Mbps.
For EcoVideo, we compute frame-wise attention entropy during the first five denoising steps and aggregate multi-step entropy signals with an exponential moving average (EMA).
Key frames are then selected based on the EMA-aggregated entropy statistics.
This design is motivated by the empirical observation that early denoising steps are particularly critical to generation quality, and EMA aggregation stabilizes entropy estimation under stochastic denoising dynamics.

\begin{table*}[t]
\caption{\textbf{Quantitative evaluation} on VBench of EcoVideo and baselines.}
\label{tab:main_table1}
\centering
\small
\setlength{\tabcolsep}{4pt}
\renewcommand{\arraystretch}{1.1}
\resizebox{\textwidth}{!}{%
\begin{tabular}{ll cccccc c c}
\toprule
Model & Method & \shortstack{Subj.\\Cons.$\uparrow$} & \shortstack{Back.\\Cons.$\uparrow$} & \shortstack{Mot.\\Smooth.$\uparrow$} & \shortstack{Aes.\\Qual.$\uparrow$} & \shortstack{Img.\\Qual.$\uparrow$} & \shortstack{Overall\\Cons.$\uparrow$} & \shortstack{VBench\\Overall$\uparrow$} & Speedup$\uparrow$ \\
\midrule

Wan2.1-14B     & --              & 0.975 & 0.981 & 0.983 & 0.661 & 0.694 & 0.259 & 0.837 & 1.00$\times$ \\
+ Wan2.1-1.3B  & HybridSD        & 0.877 & 0.966 & \textbf{0.994} & 0.435 & 0.370 & 0.152 & 0.677 & 0.89$\times$ \\
+ Wan2.1-1.3B  & EC-Diff         & 0.876 & 0.961 & 0.993 & 0.442 & 0.395 & 0.156 & 0.683 & 1.27$\times$ \\
\rowcolor{gray!15}
+ EcoVFI       & \textbf{Ours}   & \textbf{0.941} & \textbf{0.976} & 0.978 & \textbf{0.649} & \textbf{0.682} & \textbf{0.254} & \textbf{0.846} & \textbf{1.84$\times$} \\
\midrule

Wan2.2-A14B    & --              & 0.973 & 0.974 & 0.982 & 0.672 & 0.718 & 0.261 & 0.842 & 1.00$\times$ \\
+ Wan2.1-1.3B  & HybridSD        & \textbf{0.918} & 0.958 & 0.985 & 0.588 & 0.600 & 0.222 & 0.784 & 0.79$\times$ \\
+ Wan2.1-1.3B  & EC-Diff         & 0.915 & \textbf{0.965} & \textbf{0.987} & 0.598 & 0.571 & 0.229 & 0.792 & 1.02$\times$ \\
\rowcolor{gray!15}
+ EcoVFI       & \textbf{Ours}   & 0.895 & 0.956 & \textbf{0.987} & \textbf{0.660} & \textbf{0.695} & \textbf{0.261} & \textbf{0.830} & \textbf{1.59$\times$} \\
\midrule

CogVideoX-5B   & --              & 0.965 & 0.967 & 0.972 & 0.619 & 0.633 & 0.277 & 0.819 & 1.00$\times$ \\
+ CogVideoX-2B & HybridSD        & 0.872 & 0.900 & 0.967 & 0.520 & 0.567 & 0.244 & 0.725 & 0.82$\times$ \\
+ CogVideoX-2B & EC-Diff         & 0.907 & 0.908 & 0.954 & 0.486 & 0.566 & 0.228 & 0.721 & 0.98$\times$ \\
\rowcolor{gray!15}
+ EcoVFI       & \textbf{Ours}   & \textbf{0.920} & \textbf{0.960} & \textbf{0.986} & \textbf{0.560} & \textbf{0.569} & \textbf{0.260} & \textbf{0.775} & \textbf{2.03$\times$} \\
\bottomrule
\end{tabular}
}
\end{table*}

\begin{table*}[t]
\caption{\textbf{End-to-end efficiency evaluation} of EcoVideo and baselines.}
\label{tab:latency_breakdown}
\centering
\small
\setlength{\tabcolsep}{4pt}
\renewcommand{\arraystretch}{1.1}
\resizebox{\textwidth}{!}{%
\begin{tabular}{l l l c c c c c}
\toprule
Method & Cloud Model & Edge Model & Cloud Lat. (s)$\downarrow$ & Edge Lat. (s)$\downarrow$ & Comm. Vol. (MB)$\downarrow$ & Total Lat. (s)$\downarrow$ & Speedup$\uparrow$ \\
\midrule

Cloud-only & Wan2.1-14B & --          & 1996.25 & --       & 1.92           & 1996.35          & 1.00$\times$ \\
Edge-only  & --         & Wan2.1-1.3B  & --      & 3244.06  & 1.85           & 3244.06          & 0.62$\times$ \\
Hybrid-SD  & Wan2.1-14B & Wan2.1-1.3B  & 1597.24 & 648.92   & 17.23          & 2247.02          & 0.89$\times$ \\
EC-Diff    & Wan2.1-14B & Wan2.1-1.3B  & \textbf{798.96} & 778.68 & 17.23     & 1578.50          & 1.27$\times$ \\
\rowcolor{gray!15}
\textbf{Ours} & Wan2.1-14B & EcoVFI-160M     & 988.00  & \textbf{96.23} & \textbf{1.10} & \textbf{1084.29} & \textbf{1.84$\times$} \\
\addlinespace[2pt]
\midrule

Cloud-only & Wan2.2-A14B & --          & 1593.69 & --       & 2.10           & 1593.80          & 1.00$\times$ \\
Hybrid-SD  & Wan2.2-A14B & Wan2.1-1.3B  & 1194.56 & 811.43   & 17.23          & 2006.85          & 0.79$\times$ \\
EC-Diff    & Wan2.2-A14B & Wan2.1-1.3B  & \textbf{586.74} & 973.32 & 17.23     & 1560.92          & 1.02$\times$ \\
\rowcolor{gray!15}
\textbf{Ours} & Wan2.2-A14B & EcoVFI-160M     & 905.98 & \textbf{96.23} & \textbf{1.40} & \textbf{1002.28} & \textbf{1.59$\times$} \\
\addlinespace[2pt]
\midrule

Cloud-only & CogVideoX-5B & --          & 173.90  & --       & 1.40           & 173.97           & 1.00$\times$ \\
Edge-only  & --          & CogVideoX-2B  & --      & 359.25   & 1.05           & 359.25           & 0.48$\times$ \\
Hybrid-SD  & CogVideoX-5B & CogVideoX-2B  & 139.54  & 71.23    & 5.67           & 211.05           & 0.82$\times$ \\
EC-Diff    & CogVideoX-5B & CogVideoX-2B  & 69.56   & 107.94   & 5.67           & 177.78           & 0.98$\times$ \\
\rowcolor{gray!15}
\textbf{Ours} & CogVideoX-5B & EcoVFI-160M     & \textbf{63.57} & \textbf{21.95} & \textbf{0.32} & \textbf{85.54}   & \textbf{2.03$\times$} \\
\bottomrule
\end{tabular}
}
\end{table*}

\begin{figure*}[t]
    \includegraphics[width=\linewidth]{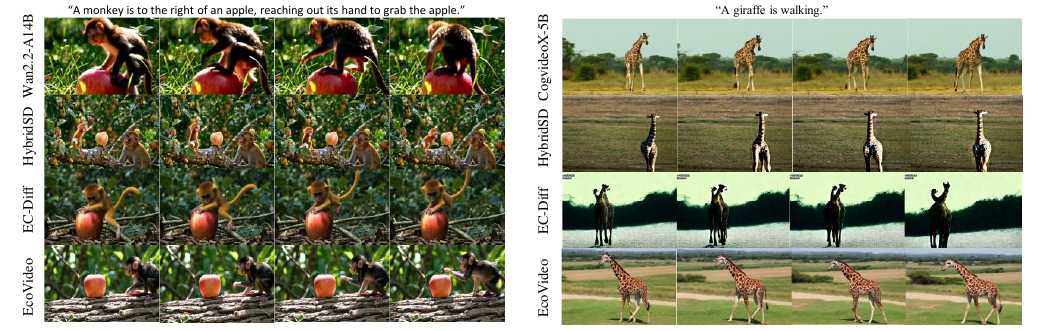}
    \caption{\textbf{Qualitative comparison} of various methods on Wan2.2 and CogVideo. Our method effectively prevents temporal artifacts such as texture flickering and detail collapse while maintaining high visual fidelity.}
    \label{fig:qual_comp}
\end{figure*}

\subsection{Main Result}

\noindent\textbf{Quality Evaluation.}
To quantitatively evaluate generation quality, we generate videos using the VBench prompt set and compare EcoVideo with step-splitting cloud--edge baselines, namely HybridSD and EC-Diff, under the same cloud-side large model. Table~\ref{tab:main_table1} summarizes VBench results across six representative dimensions. Overall, EcoVideo consistently outperforms existing collaborative baselines across all three model families in terms of VBench Overall score, yielding an average improvement of 0.084 over the best step-wise collaboration baseline. Beyond the overall metric, EcoVideo also delivers consistent gains in perceptual quality, including aesthetic quality, image quality, and overall consistency, while maintaining competitive temporal smoothness. These results indicate that fine-grained frame-wise cloud-edge orchestration is more effective than step-wise collaboration in preserving visual fidelity and temporal coherence during accelerated video generation.

\noindent\textbf{Efficiency Evaluation.}
We evaluate end-to-end efficiency in the cloud--edge system and further decompose the overhead into cloud latency, edge latency, and communication volume at Table~\ref{tab:latency_breakdown}.
Compared to HybridSD and EC-Diff with fixed splitting strategies, EcoVideo significantly reduces both communication and edge overhead.
Under the Wan2.1-14B setting, the communication volume drops from 17.23~MB for both HybridSD and EC-Diff to 1.10~MB.
The edge latency decreases from 648.92~s for HybridSD and 778.68~s for EC-Diff to 96.23~s.
Under the CogVideoX-5B setting, EcoVideo further reduces communication to 0.32~MB, which is below 5.67~MB required by the step-splitting baselines.
It also reduces the edge latency from 71.23~s for HybridSD and 107.94~s for EC-Diff to 21.95~s, reflecting stronger robustness to bandwidth fluctuations.

\noindent\textbf{Qualitative Visualizations.}
Figure~\ref{fig:qual_comp} presents qualitative comparisons across different model pairings.
Compared to HybridSD and EC-Diff, EcoVideo substantially mitigates typical temporal artifacts such as texture flickering and detail collapse, maintaining better visual consistency while achieving a higher acceleration ratio.
For example, when generating a video of a monkey to the right of an apple, reaching out its hand to grab the apple, or a giraffe walking, the step decoupling baseline may produce ghosting and duplicated outlines in certain frames, breaking temporal consistency.
In contrast, EcoVideo maintains consistent content across frames and ensures smooth temporal transitions, preserving semantic integrity throughout the entire sequence.

\subsection{Ablation Study}

\begin{figure}[t]
    \centering
    \begin{minipage}[t]{0.49\textwidth}
        \vspace{-10pt}
        \centering
        \setlength{\tabcolsep}{1pt}
        \captionof{table}{\textbf{Ablation on Key Module and Warm-up.}}
        \vspace{6pt} 
        \label{tab:breakdown_table}
        \scalebox{0.85}{
        \begin{tabular}{lccc}
            \toprule
            \textbf{Method} & \textbf{VBench} & \textbf{Latency(s)} & \textbf{Speedup} \\
            \midrule
            Wan2.1-14B & 0.837 & 1996.35 & 1.00$\times$ \\
            EcoVideo & 0.845 & 1084.29 & 1.84$\times$ \\
            w/o Entropy & 0.832 & 1049.73 & 1.90$\times$ \\
            w/o EcoVFI & 0.835 & 988.82 & 2.02$\times$ \\
            w/ naive interp. & 0.835 & 1019.33 & 1.96$\times$ \\
            \midrule
            Warmup 2\% & 0.834 & 938.17 & 2.13$\times$ \\
            Warmup 20\% & 0.842 & 1180.13 & 1.69$\times$ \\
            Warmup 30\% & 0.847 & 1336.45 & 1.49$\times$ \\
            \bottomrule
        \end{tabular}
        }
    \end{minipage}
    \hfill
    \begin{minipage}[t]{0.49\textwidth}
        \centering
        \vspace{0pt}
        \includegraphics[width=\textwidth]{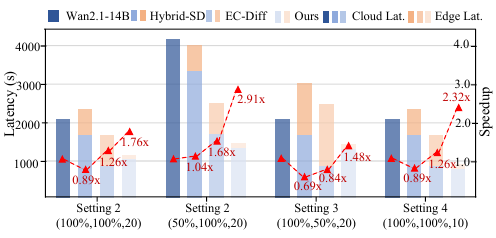}
        \captionof{figure}{\textbf{Cloud-edge breakdown analysis} of ours and baselines. Each setting is (Cloud GPU, Edge GPU, Bandwidth).}
        \label{fig:breakdown}
    \end{minipage}
\end{figure}

\noindent\textbf{Ablation on Key Modules.}
We ablate the key components of EcoVideo to identify the source of its quality--efficiency gains, as shown in Table~\ref{tab:breakdown_table}.
Replacing entropy-based keyframe selection with uniform keyframes without entropy improves speedup from 1.84$\times$ to 1.90$\times$, but drops VBench from 0.845 to 0.832, showing that entropy-based selection better preserves informative temporal changes.
Removing EcoVFI and generating only keyframe videos without EcoVFI further improves speedup to 2.02$\times$, but lowers VBench to 0.835, indicating that edge-side frame reconstruction is necessary for temporal completeness.
Using the original EDEN interpolation with naive interpolation also yields a lower VBench of 0.835, suggesting that interpolation alone cannot account for the gains.
The full EcoVideo achieves the best trade-off, confirming that the improvement comes from the combination of entropy-based keyframe selection, model collaboration, and information-reuse-based interpolation.

\noindent\textbf{Ablation on Warm-up Initialization.}
We ablate the warm-up length for entropy-EMA stabilization by finalizing entropy statistics at different warm-up ratios as shown in Table~\ref{tab:breakdown_table}.
A shorter warm-up reduces computation but provides less stable entropy estimates. Using 2\% warm-up achieves the highest speedup of 2.13$\times$ but lowers VBench to 0.834.
Increasing the warm-up ratio improves entropy reliability and VBench. The score reaches 0.842 at 20\% and 0.847 at 30\%. However, latency increases and speedup drops to 1.69$\times$ and 1.49$\times$.
We adopt 10\% warm-up as the default setting since it provides a better balance between entropy estimation reliability and end-to-end efficiency.

\begin{figure*}[t]
    \includegraphics[width=\linewidth]{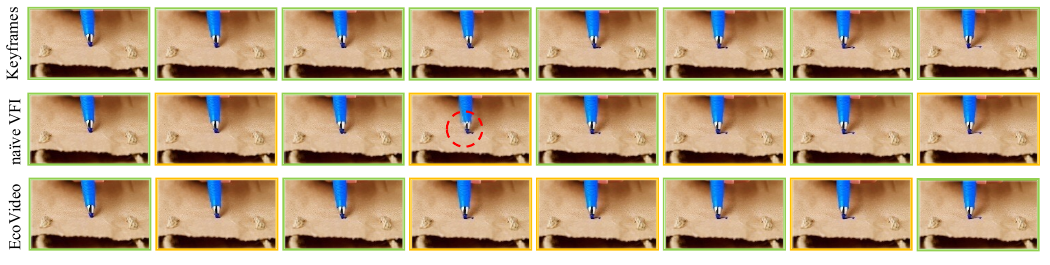}
    \caption{Keyframe visualization of EcoVideo on Wan2.1-14B. Green borders indicate keyframes, and yellow borders indicate interpolated non-keyframes.}
    \label{fig:vis_vfi}
\end{figure*}

\noindent\textbf{Ablation on the Interpolation Module.}
To verify the necessity of EcoVFI, we further qualitatively compare different intermediate-frame completion strategies in Figure~\ref{fig:vis_vfi} while keeping key frames fixed, including: (i) cloud-only large-model generation, (ii) a naive interpolation strategy with original EDEN\cite{zhang2025eden}, and (iii) EcoVFI interpolation.
Naive interpolation tends to produce ghosting, texture breaks, and motion discontinuities in fast-motion regions and fine-texture areas.
In contrast, EcoVFI significantly improves structural consistency and motion coherence for intermediate frames. Visually, it is closer to the frame-by-frame reference generated by the large model while preserving EcoVideo's acceleration advantage.
This ablation shows that high-quality interpolation is essential for temporal coherence under sparse keyframe generation.

\noindent\textbf{Ablation on System-Aware Dynamic Optimization.} We evaluate EcoVideo under representative resource settings, as shown in Fig.~\ref{fig:breakdown}. Each setting is (Cloud GPU, Edge GPU, Bandwidth): default, cloud contention, edge contention, and bandwidth reduction, respectively. Fixed step-splitting baselines are sensitive to bottleneck shifts: when cloud compute is contended, HybridSD and EC-Diff drop to 1.04$\times$ and 1.68$\times$, while EcoVideo achieves 2.91$\times$ speedup. When edge compute is contended, HybridSD and EC-Diff fall below cloud-only throughput with 0.69$\times$ and 0.84$\times$, whereas EcoVideo remains faster at 1.48$\times$. Under reduced bandwidth, EcoVideo reaches 2.32$\times$ speedup, outperforming HybridSD and EC-Diff at 0.89$\times$ and 1.26$\times$, mainly due to its much lower transmitted volume of 1.0--1.2~MB versus 17.23~MB. These results show that system-aware adaptation improves robustness under dynamic compute and bandwidth conditions.
\section{Related Work}

\noindent\textbf{Video Generation Paradigm.}
Video generation has evolved from explicit distribution modeling represented by Generative Adversarial Networks (GANs)\cite{creswell2018generative} and Variational Autoencoders (VAEs)\cite{kingma2013auto} to diffusion-based iterative denoising\cite{ho2020denoising}, and the diffusion Transformer-based scaling of models\cite{vaswani2017attention,peebles2023scalable} further strengthens this paradigm.
With the continued growth of data and model size, DiT-style models\cite{peebles2023scalable} such as Sora\cite{brooks2024video} and Wan\cite{wan2025wan,yang2024cogvideox,kong2024hunyuanvideo,hacohen2024ltx} demonstrate longer duration, higher fidelity, and more coherent video generation.
Nevertheless, these models still follow a step-by-step, full-frame denoising inference procedure, making the computation cost grow rapidly with both the number of denoising steps and the video length, which results in high latency and cost and becomes a key bottleneck for large-scale online services.

\noindent\textbf{Efficient Video Generation.}
Improving the service efficiency of video generation is commonly achieved via distributed parallelism\cite{fang2024pipefusion,li2024distrifusion,xiang2025macro}, caching reuse\cite{liu2025timestep,zou2024accelerating,wei2025mixcache}, and sparse attention\cite{xi2025sparse,zhang2025fast,luo2026attention} to accelerate video diffusion inference.
Recently, a line of work has reduced computation by decoupling the generation process along axes such as denoising steps\cite{jin2024pyramidal,zhang2025flashvideo}, resolution\cite{skorokhodov2024hierarchical,zhang2025flashvideo}, and frame\cite{ma2025tempomaster,wang2025keyvid,yin2023nuwa}.
Some efforts further deploy such decoupled paradigms to cloud--edge settings: HybridSD\cite{yan2024hybrid} assigns early diffusion steps to a cloud large model for semantic planning and later steps to an edge small model for refinement to reduce cloud cost, while EC-Diff\cite{xie2025ec} further reduces cloud inference frequency via noise-gradient approximation and searches the cloud--edge switching point to trade off quality and speed.
Unlike step decoupling methods, EcoVideo further improves temporal consistency and cloud--edge deployment efficiency by decoupling generation at the frame level. It is also orthogonal to existing acceleration techniques such as parallelism, caching, and sparse attention.
\section{Limitations and Future Directions} \label{sec:limitation}
EcoVideo targets cloud--edge video generation and is complementary to cloud-only acceleration methods such as parallelism, caching reuse, and sparse attention, which can further accelerate cloud-selected keyframes. Its reliance on edge-side interpolation may struggle under extremely fast motion, heavy occlusion, or abrupt scene changes, where more keyframes or stronger interpolation models may be required. Future work will explore joint optimization with cloud-side acceleration, uncertainty-aware keyframe insertion, and long-running online adaptation on real-world network traces and heterogeneous edge devices.

\section{Conclusion}
\label{sec:conclusion}
In this work, we present \textbf{EcoVideo}, an entropy-orchestrated cloud--edge framework for efficient and stable diffusion video generation under dynamic bandwidth and edge compute. 
    EcoVideo uses early-stage attention entropy to estimate inter-step information density and orchestrate generation at the frame level: high-entropy frames are denoised on the large cloud model, while low-entropy frames are completed on a small edge model. 
    EcoVideo improves temporal coherence while reducing cloud computation and communication. Experiments show favorable quality-efficiency trade-offs and effective mitigation of common temporal artifacts. 
Our results highlight entropy-driven orchestration as a practical direction for robust video generation in cloud-edge systems.


%
%

\bibliographystyle{_sty/splncs04}
\bibliography{_ref/2_VGM}


\end{document}